\documentclass[runningheads]{llncs}
\usepackage{graphicx}
\usepackage[title,toc,titletoc,header]{appendix}
\usepackage{amsmath}
\usepackage{amssymb}
\usepackage{subfigure}
\usepackage{booktabs}
\usepackage{array}
\usepackage{multirow}
\usepackage{kotex}
\usepackage{xcolor}
\newcolumntype{L}[1]{>{\raggedright\let\newline\\\arraybackslash\hspace{0pt}}m{#1}}
\newcolumntype{C}[1]{>{\centering\let\newline\\\arraybackslash\hspace{0pt}}m{#1}}
\newcolumntype{R}[1]{>{\raggedleft\let\newline\\\arraybackslash\hspace{0pt}}m{#1}}

\newcommand\Tstrut{\rule{0pt}{2.5ex}}         

\begin{document}
\title{Drug-disease Graph: Predicting Adverse Drug Reaction Signals via Graph Neural Network with Clinical Data}

\titlerunning{Drug-disease Graph: Predicting ADR Signals via GNN with Clinical Data}
%
\author{Heeyoung Kwak\inst{1}
\and Minwoo Lee\inst{1} \and
Seunghyun Yoon\inst{1} \and
Jooyoung Chang\inst{2} \and
Sangmin Park\inst{2,3} \and
Kyomin Jung\inst{1,}$^*$}

\authorrunning{H. Kwak et al.}
%
\institute{Department of Electrical and Computer Engineering, \\Seoul National University, Seoul, Korea\\
\email{\{hykwak88,minwoolee,mysmilesh,kjung\}@snu.ac.kr}
\and
Department of Biomedical Sciences, Seoul National University, Seoul, Korea\\
\email{\{joomyjoo,smpark.snuh\}@gmail.com}
\and
Department of Family Medicine, Seoul National University Hospital, Seoul, Korea}

\maketitle              
\begin{abstract}
Adverse Drug Reaction (ADR) is a significant public health concern world-wide.
Numerous graph-based methods have been applied to biomedical graphs for predicting ADRs in pre-marketing phases.
ADR detection in post-market surveillance is no less important than pre-marketing assessment, and ADR detection with large-scale clinical data have attracted much attention in recent years. However, there are not many studies considering graph structures from clinical data for detecting an ADR signal, which is a pair of a prescription and a diagnosis that might be a potential ADR. In this study, we develop a novel graph-based framework for ADR signal detection using healthcare claims data. We construct a Drug-disease graph with nodes representing the medical codes. The edges are given as the relationships between two codes, computed using the data. We apply Graph Neural Network to predict ADR signals, using labels from the Side Effect Resource database. The model shows improved AUROC and AUPRC performance of 0.795 and 0.775, compared to other algorithms, showing that it successfully learns node representations expressive of those relationships. Furthermore, our model predicts ADR pairs that do not exist in the established ADR database, showing its capability to supplement the ADR database.




\end{abstract}

\section{Introduction}
\label{sec:introduction}

An adverse drug reaction (ADR) is considered to be one of the significant causes of morbidity and mortality, estimated to be the fourth to sixth highest cause of death in the United States~\cite{lazarou1998incidence}. Most ADR detection research has been aimed to predict ADRs in pre-marketing phases, using biomedical information sources such as chemical structures, protein targets, and therapeutic indications. Especially, studies using graph-structured data
have demonstrated the superiority of modeling biomedical interactions as graphs. 
Nevertheless, capturing potential ADRs from the entire population in post-marketing phases is also essential to fully establish the ADR profiles~\cite{jeong2018machine}. The potential causal relationship between an adverse event and a drug is called a `signal' when the relation is previously unknown or incompletely documented. Traditional ADR signal detection research in post-marketing phases mainly counts on a spontaneous and voluntary reporting system that collects spontaneous reports of suspected drug-related events, such as the WHO Uppsala Monitoring Center~\cite{bate2002data,hauben2009decision,hochberg2009time}. 
However, the spontaneous reporting system has inherent limitations such as underreporting~\cite{hazell2006under,sarker2015utilizing}, selective reporting~\cite{hochberg2009time}, and the lack of drug usage data. Therefore, recent studies have attempted algorithmic approaches to detect ADR signals on large clinical databases such as electronic health records (EHR) and healthcare claims data~\cite{jeong2018machine,park2011novel}.
Many of these studies apply basic machine learning techniques such as random forest, support vector machines, and neural networks. However, fewer studies are using graph-based approaches on the clinical databases in the field of post-marketing ADR signal detection. 
Due to the complex polypharmacy and multiple relations among drugs and diseases, we expect that graph structure can provide insights to potential ADRs, which may not otherwise be apparent using disconnected structures.

\begin{figure}[t]
\centering
\includegraphics[width=\columnwidth]{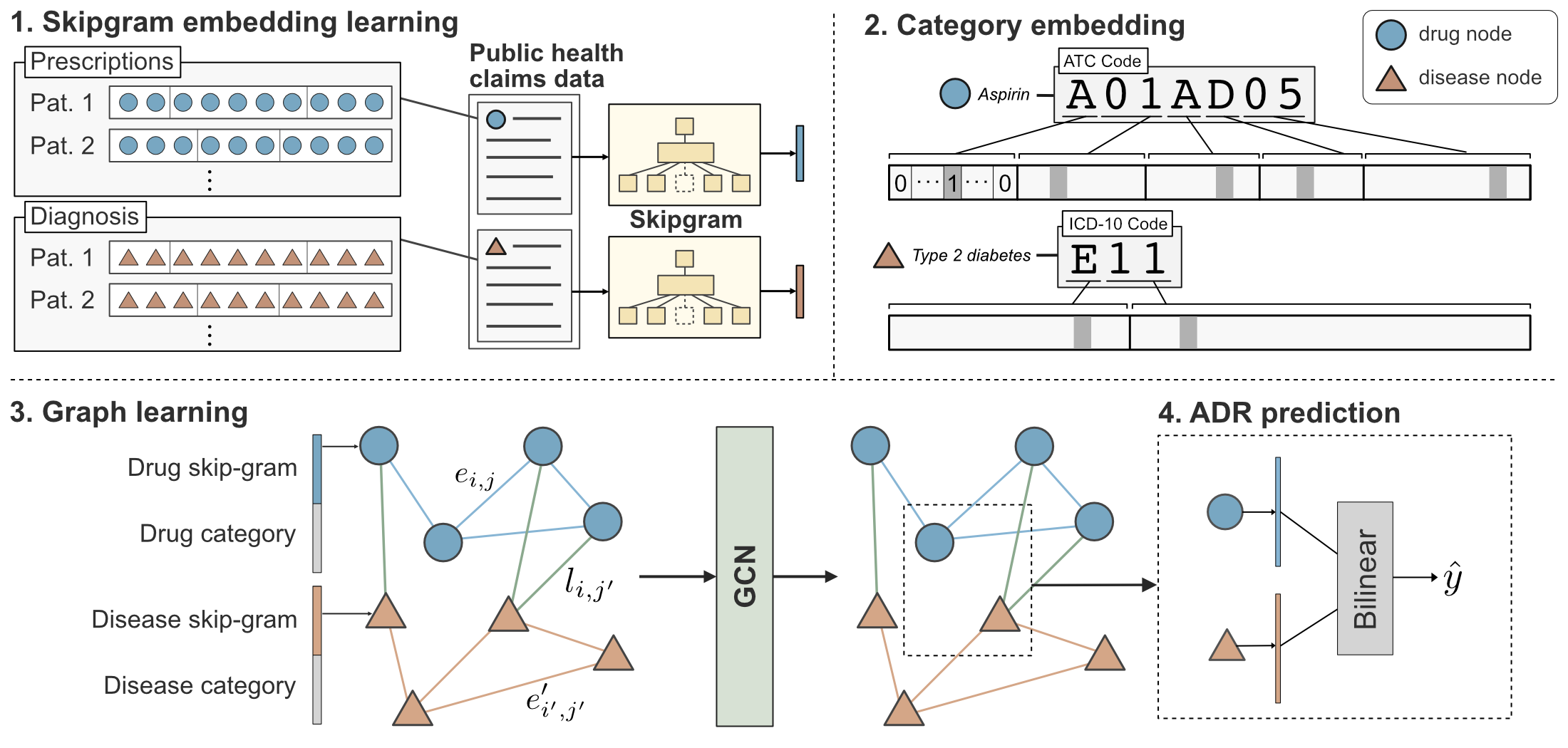}
\caption{
Overview of the proposed ADR detection task pipeline. 
Drug and disease embeddings are learned using Skip-gram model on clinical data. 
Categorical embedding is extracted from the identifier codes of drug and diseases.
Skip-gram and categorical embeddings are jointly used as input of node feature in the heterogeneous graph, and the graph is passed through GNN for ADR prediction.
}
\label{fig:topology}
\end{figure}
In this study, we develop a novel graph-based framework for ADR signal detection using healthcare claims data to construct a Drug-disease graph. Specifically, we use National Health Insurance Service-National Sample Cohort (NHIS-NSC), the 12-year healthcare claims data that covers medical histories for one million population~\cite{lee2016cohort}.
The constructed graph is a heterogeneous graph with drug and disease nodes, as it is depicted in Figure~\ref{fig:topology}. The nodes represent the medicine prescription codes and disease diagnosis codes derived from the healthcare claims data. To represent the relations among these codes, we define edge weights using information from the data. For example, \textit{l2}-distance between two node embeddings, which are learned from the data, is used to define the drug-drug and disease-disease edge weights. Also, the conditional probability computed on the data is used for the drug-disease relationship. As Graph Neural Network (GNN) models have been demonstrated \cite{kipf2016semi,velivckovic2017graph} their power to solve many tasks with graph-structured data, showing state-of-the-art performances, we use GNN-based approach for ADR detection. We verify that GNNs can learn node representations that are indicative of various relations between drugs and diseases. Then our model makes a prediction on whether a drug node and a disease node will have an ADR relation based on the learned node representations.

To evaluate the performance of the proposed approaches, we conduct experiments with the newly generated dataset using the side effect resource database (SIDER).
The empirical results demonstrate the superiority of our proposed model, which outperforms other alternative machine learning algorithms with a significant margin in terms of the area under the receiver operating characteristic (AUROC) score and the area under the precision-recall curve (AUPRC) score.
Furthermore, our method unveils ADR candidates that are examined to be very useful information to the medical community. Our model uses only simple data processing and well-established medical terminologies. Therefore, our work does not demand case-by-case feature engineering that requires expertise.

\section{Related Works}
\label{sec:related_works}
There have been numerous studies on ADR prediction in pre-marketing phases, attempting graph-based approaches on biomedical information sources~\cite{liu2012large,pauwels2011predicting,su2018network,yue2019graph}. These studies predicted potential side-effects of drug candidate molecules based on their chemical structures~\cite{pauwels2011predicting} and additional biological properties~\cite{liu2012large}. Although such studies may play important roles in preventing ADRs in pre-marketing phases, capturing potential ADRs in real-world use cases has been considered very important.

A spontaneous and voluntary reporting system has been an important data source of the real world drug usages. Most of the traditional ADR signal detection research used voluntary reporting systems with disproportionality analysis (DA), which measures disproportionality of observed drug-adverse event pairs existing in data and the null expectations~\cite{bate2002data,hauben2009decision,hochberg2009time}. Recently, large-scale clinical databases such as EHR (Electronic Health Records) or healthcare claims data have gained popularity as an alternative or additive data source in ADR signal detection research.
Much of the studies applied machine learning techniques such as support vector machine (SVM), random forest (RF), logistic regression (LR) and other statistical machine learning methods to model the decision boundary to detect ADR in post-marketing phases~\cite{jeong2018machine,karlsson2013predicting,liu2012comparative,park2011novel,yoon2012detection}.

More recently, researchers explored neural network-based models over clinical databases.
Shang et al.~\cite{shang2019gamenet} combined graph structure with the memory network to recommend a personalized medication. The longitudinal electronic health records and drug-drug interaction information were embedded as a separate graph to be jointly considered for the recommendation. There also exists research for the recommendation, but the architectures are limited to the single use of instance symptoms~\cite{wang2017safe,zhang2017leap}, or patient history~\cite{le2018dual}.
However, none of these research explored graph neural network model for predicting the ADR reactions in the post-marketing phase.

\section{The Proposed Framework}
In this section, we formulate our problem and describe how we apply graph structures for the task. We also present the process of training and prediction.

\subsection{Problem Formulation}
The task is to predict the potential causal relationship between a given drug and a disease pair, which represents the prescription code and the diagnosis code in clinical data. To consider the various relationships between drugs and diseases, we convert our clinical data into a novel graph structure that consists of drug and disease nodes. The node representations and the edge weights are given according to the information retrieved from the clinical data NHIS-NSC in this study.
We first learn a node embedding that reflects the temporal proximity between homogeneous nodes, i.e., drug-drug and disease-disease node pairs. In order to model the proximity between two codes, we form drug/disease sequences from patients' records.

After the drug-disease graph is constructed, we build a GNN model that predicts the signal of side effects between any pairs of drug and disease. The side effect labels, which are taken from the SIDER database, are given to a subset of drug-disease pairs in graph $G$.
We define the label function $l \colon V_{\text{drug}}^{\text{SIDER}} \times V_{\text{dis}}^{\text{SIDER}} \to \{0\,,1\}$ as follows:
\begin{equation}
\begin{aligned}
& l(v,w)=
\left\{
    \begin{array}{ll}
        1  &~~~~~~~\mbox{if } (v,w) \in E^{\text{SIDER}}, \\
        0 &~~~~~~~\mbox{otherwise,}
    \end{array}
\right.
\end{aligned}
\label{eq:topology_drug_sider}
\end{equation}
where $V_{\text{drug}}^{\text{SIDER}}$ and $V_{\text{dis}}^{\text{SIDER}}$ are the subsets of $V_{\text{drug}}$ and $V_{\text{dis}}$ registered in the SIDER database respectively, and $E^{\text{SIDER}}$ is the set of drug-disease pairs that are known to have side effect relation according to the SIDER database.
\subsection{Code Embedding Learning with Skip-gram Model}
\label{ssec:embedding-skip-gram}
Most large-scale clinical databases including NHIS-NSC, are collected in the form of longitudinal visit records of the patients. In this section, we explain how we process the patient's longitudinal records as sequential data and apply skip-gram model to learn the code embeddings.\\\\ 
\noindent{\textbf{Definition 1 (Drug/Disease Sequence)}} In the patient's longitudinal records, each patient can be treated as a sequence of hospital visits $\{v_1^{(n)}, v_2^{(n)},...,v_{T_n}^{(n)}\}$ where $n$ represents each patient in the data, and $T_n$ is the total number of visits of the patient.
The $i^{th}$ visit can be denoted as $v_i^{(n)}\,{=}\,\{{\mathcal{P}_i}^{(n)},{\mathcal{D}_i}^{(n)}\}$ where ${\mathcal{P}_i}^{(n)}$ is the set of prescribed codes and ${\mathcal{D}_i}^{(n)}$ is the set of diagnosed codes in the $i^{th}$ visit. Within a set of codes, codes are listed in arbitrary order. The size of each set is variable since the number of prescribed/diagnosed codes varies from visit to visit. With these sets of codes, we form a drug sequence ${{\textbf{Seq}}_{drug}}^{(n)}$ and a disease sequence ${{\textbf{Seq}}_{disease}}^{(n)}$ of $n^{th}$ patient by listing each of the codes in a temporal order, as it is described below (Here, we leave out the symbol $n$):
\begin{equation}
\begin{aligned}
&{\textbf{Seq}}_{drug} = \{p_1,p_2\,,...\,,p_{T_p}\}\,,\, p_x \in {\mathcal{P}}_i, \\ 
&{\textbf{Seq}}_{disease} = \{d_1,d_2\,,...\,,d_{T_d}\}\,,\, d_y \in {\mathcal{D}}_i, 
\end{aligned}
\label{eq:definition_1}
\end{equation}
where $p_x \in \mathbb{R}^{V_p}$ and $d_y \in \mathbb{R}^{V_d} $ are the one-hot vectors representing each of the medical codes in the sequences. $V_p$ and $V_d$ are the vocabulary size of the whole prescription and diagnosis codes within the data, respectively. $T_p$ and $T_d$ represent the total number of prescription/diagnosis codes of the patient's record. In this way, we can build a corpus consisting of ${\textbf{Seq}}_{drug}$ or ${\textbf{Seq}}_{disease}$, in which the proximity-based code embedding learning can be implemented.

We use Skip-gram \cite{mikolov2013distributed} model to learn the latent representation of medical codes in our data, in a way that captures the temporal proximity between them. With ${\textbf{Seq}}_{drug}$ or ${\textbf{Seq}}_{disease}$, we use the context window size of 16, meaning 16 codes behind and 16 codes ahead, and apply the Skip-gram learning with negative sampling scheme. As a result, we project both diagnosis codes and prescription codes into the separate lower-dimensional spaces, where codes are embedded close to one another that are in close proximity to them. The trained Skip-gram vectors are then used as the proximity-based code embeddings.

\subsection{Drug-disease Graph Construction}
\label{ssec:drug-disease-graph}
Here, we describe how we construct our unique Drug-disease graph from NHIS-NSC. In Definition 2, we explain the concept of the Drug-disease graph. Then, we explain the node representations and edge connections.\\

\noindent{\textbf{Definition 2 (Drug-disease Graph)}} We construct a single heterogeneous graph $\mathcal{G}\,{=}\,(\mathcal{V}, \mathcal{E})$ consisting of drug and disease nodes, where $\mathcal{V}\,{=}\,\mathcal{V}_{\text{drug}} \cup \mathcal{V}_{\text{dis}}$ is the union of drug and disease nodes, and $\mathcal{E}\,{=}\,\mathcal{E}_{\text{drug}} \cup \mathcal{E}_{\text{dis}} \cup \mathcal{E}_{\text{inter}}$ is the union of homogeneous edges $\mathcal{E}_{\text{drug}}$ and $\mathcal{E}_{\text{dis}}$ (i.e. consisting of same type of nodes) and heterogeneous edges $\mathcal{E}_{\text{inter}}$  (i.e. consisting of different types of nodes). \\\\
To represent $\textbf{v}_{\text{drug}} \in \mathcal{V}_{\text{drug}}$ and $\textbf{v}_{\text{dis}} \in \mathcal{V}_{\text{dis}}$, we jointly use proximity-based node representation along with category-based node representation. Proximity-based node representation is obtained by initial Skip-gram code embedding as in section~\ref{ssec:embedding-skip-gram}. We denote a proximity-based drug node as $\textbf{v}'_{\text{drug}}$ and a disease node as $\textbf{v}'_{\text{dis}}$. Category-based node representation is designed to represent the categorical information of medical codes. We utilize the hierarchical structure of categorical codes (i.e. ATC and ICD-10 codes) by adopting the one-hot vector format. Since there are multiple categories for each code, the category-based node representation is shown as a concatenation of one-hot vectors, thus, a multi-hot vector. Finally, the initial node representation of the Drug-disease graph are represented as the concatenation of the proximity-based node embeddings and the category-based node embeddings. Following are the definitions for the drug and disease node representations.\\

\noindent{\textbf{Definition 3 (Node Representations)}}
\vspace*{1mm}
\begin{equation}
\begin{aligned}
&\textbf{v}''_{drug} = \{\textbf{v}_{drug}^1 || \textbf{v}_{drug}^2 ||\textbf{v}_{drug}^3 || \textbf{v}_{drug}^4||\textbf{v}_{drug}^5\}, \\
&\textbf{v}''_{dis} = \{\textbf{v}_{dis}^1 || \textbf{v}_{dis}^2\}, \\
&\textbf{v}_{drug} = \{\textbf{v}'_{drug}\, || \,\textbf{v}''_{drug}\},\\
&\textbf{v}_{dis} = \{\textbf{v}'_{dis} \,||\, \textbf{v}''_{dis}\}, 
\end{aligned}
\label{eq:categorical-node}
\end{equation}
where $\textbf{v}''_{drug}$ is a category-based drug node, $\textbf{v}''_{dis}$ is a category-based disease node, $\textbf{v}_{drug}$ is an initial drug node, $\textbf{v}_{dis}$ is an initial disease node, and $||$ is a vector concatenation function.
Each $\textbf{v}^i_{drug}$ represents the each level in the ATC code structure and $\textbf{v}''_{drug} \,{\in}\,\mathbb{R}^{104}$. Because the ATC code structure is represented in 5 levels, a drug node vector is also represented as the concatenation of 5 one-hot vectors. Similarly, each $\textbf{v}_{dis}^i$ represents each of the first two levels in the ICD-10 code structure and $\textbf{v}''_{dis} \,{\in}\,\mathbb{R}^{126}$. We only use two classification levels of the ICD-10 code structure, therefore, the disease node vector is represented as the concatenation of 2 one-hot vectors.

For homogeneous edges like $\mathcal{E}_{\text{drug}}$ and $\mathcal{E}_{\text{dis}}$, we view the relationships between homogeneous nodes as the temporal proximity of two entities, meaning that two nodes are likely to be close together in the records. Therefore, using the proximity-based node embeddings, we compute \textit{l2}-distance between two node embeddings to estimate the temporal proximity. For heterogeneous edges, which are the edges connecting drug nodes and disease nodes, are given as the conditional probability of drug prescription given the diagnosed disease. The definitions of the two types of edges are given as follows:\\

\noindent{\textbf{Definition 4 (Homogeneous Edges)}}
For any node $i,j \in \mathcal{V}_{\text{drug}}$ (or $\mathcal{V}_{\text{dis}}$), the edge weight $w_{ij}$ between two nodes are defined using Gaussian weighting function as follows: 
\begin{equation}
\begin{aligned}
& w_{ij}=
\left\{
\begin{array}{ll}
	exp(-\frac{{\lVert \textbf{v}'_i - \textbf{v}'_j \rVert}^2}{2\theta^2}) &~~~~~~~~\mbox{if }   \lVert \textbf{v}'_i - \textbf{v}'_j \rVert \leq threshold, \\
	0 &~~~~~~~~\mbox{otherwise,}
\end{array}
\right.
\end{aligned}
\label{eq:homo_edge}
\end{equation}
for some parameters \textit{threshold} and $\theta$. $\textbf{v}'_i$ and $\textbf{v}'_j$ are the proximity-based node embeddings of two nodes $i$ and $j$. Later, we additionally use edge-forming thresholds to control the sparsity of the graph.\\

\noindent{\textbf{Definition 5 (Heterogeneous Edges)}}
For any drug node $i \in \mathcal{V}_{\text{drug}}$ and any disease node $j \in \mathcal{V}_{\text{dis}}$, the edge weight $w_{ij}$ between two nodes are given as: 
\begin{equation}
\begin{aligned}
& w_{ij}=\frac{n_{ij}}{n_j},
\end{aligned}
\label{eq:topology}
\end{equation}
where $n_{ij}$ is number of patients' histories in the NHIS-NSC database that is recorded with a diagnosis $j$ and a prescription $i$ in tandem. $n_j$ is the number of patients' histories with the diagnosis $j$. 

\subsection{A GNN-based Method for Learning Graph Structure}

We aggregate neighborhood information of each drug/disease node from the constructed graph using the Graph Neural Network (GNN) framework. In each layer of GNN, the weighted sum of neighboring node features in the previous layer is computed to serve as the node features (after applying a RELU nonlinearity $\sigma$) as follows:
\begin{equation}
\begin{aligned}
& \mathbf{z_i}^{(l+1)}=\sigma(\sum_{j\in \mathcal{N}(i)}{{\alpha_{ij}^{(l)}} W \mathbf{z_j}^{(l)}}), \\
\end{aligned}
\label{eq:gcn_framework}
\end{equation}
where $\mathcal{N}(i)$ denotes the set of neighbors of $i^{th}$ node, $\mathbf{z_i}^{(l)}$ denotes feature vector of $i^{th}$ node at $l^{th}$ layer, $W$ denotes a learnable weight matrix and $\alpha_{ij}^{(l)}$ denotes the normalized edge weight between $i^{th}$ and $j^{th}$ nodes at the $l^{th}$ layer. 
In the first layer, the initial drug/disease node representations are each passed through a nonlinear projection function to match their dimensions.

We use two weighting schemes for $\alpha_{ij}^{(l)}$. The first variant follows the definition in \cite{kipf2016semi}, and the weight is defined as follows: 
\begin{equation}
\begin{aligned}
& \alpha_{ij} = \frac{w_{ij}}{\sqrt{d_i d_j}},
\end{aligned}
\label{eq:gcn}
\end{equation}
where $d_i$ and $d_j$ are the degree of nodes $i$ and $j$ respectively, and $w_{ij}$ are the edge weights defined in section~\ref{ssec:drug-disease-graph}. The weights are fixed  for all layers. The second weighting scheme instead learns the weighting scheme using attention mechanism \cite{velivckovic2017graph} as follows:
\begin{equation}
\begin{aligned}
& \alpha_{ij}^{(l)}=\frac{\exp(g(\mathbf{z_i}^{(l)},\mathbf{z_j}^{(l)}))}{\sum_{k\in\mathcal{N}(i)}{\exp(g(\mathbf{z_i}^{(l)},\mathbf{z_k}^{(l)}))}},
\end{aligned}
\label{eq:gat}
\end{equation}
where $g$ is a single fully-connected layer with LeakyReLU nonlinearity that takes a pair of node features as input. In the rest of this paper, we call the network with the first weighting scheme as \textbf{GCN} and the network with the second scheme as \textbf{GAT}.

We predict the ADR signal of a drug-disease pair using the learned embeddings from the GNN model with a single bilinear layer as follows:
\begin{equation}
\begin{aligned}
& \hat{y}_{ij}=\sigma({\mathbf{z_i}^{(L)}}W_p \mathbf{z_j}^{(L)}+b),
\end{aligned}
\label{eq:bilinear}
\end{equation}
where $W_p$, $b$ are the learnable weights, and $v_i^{(L)}$, $v_j^{(L)}$ are the node features of drug node $i$ and disease node $j$ at the last GNN layer. The whole model is trained by minimizing the cross-entropy loss. 

\section{Experiments}
\subsection{Data preprocessing}
As we get the labels from the SIDER database and the edge weight from the NHIS-NSC database, we retrieve the drug and disease nodes over the joint set of two databases. The resulting dataset is composed of 607 drugs and 556 diseases, and the number of positive samples, indicating the drug-side effect relationships, are 28,746 pairs. A negative sample is defined as a combination of drugs and diseases over the dataset, where the known 28,746 positive samples are excluded. We randomly select negative samples, setting the size of negative samples same as the size of positive samples.

\subsection{Experimental Settings}
Since we extract those combinations from the SIDER database, it is plausible to believe that they have not been reported as ADRs. Although the labels are only given to the drug-disease pairs over the joint set of two databases, we make use of all the drugs and diseases in NHIS-NSC as graph nodes to utilize the relations among the drugs and diseases.

\begin{table}[t]
\centering
\small
\caption{
Summary statistics of the constructed graph and datasets
}
\begin{tabular}{L{0.65\columnwidth}C{0.15\columnwidth}C{0.15\columnwidth}}
\toprule
\textbf{Edge-forming threshold}\Tstrut & \textbf{Low}\Tstrut & \textbf{High}\Tstrut \\
\midrule
\# Drug nodes & \multicolumn{2}{c}{1,201} \\
\# Labeled drug-dis pairs in train & \multicolumn{2}{c}{37,016} \\
\# Labeled drug-dis pairs in test & \multicolumn{2}{c}{6,092}  \\
\# Disease nodes          & \multicolumn{2}{c}{10,117}  \\ 
\# Drug2drug-Edges        & {19,918}    &  {7,199} \\
\# Drug2dis-Edges         & \multicolumn{2}{c}{1,306} \\
\# Dis2dis-Edge           & \multicolumn{2}{c}{401,801} \\ 
\bottomrule
\end{tabular}
\label{table:statistics}
\end{table}

To predict the link between the drugs and diseases, we split drug-disease pairs from the ADR dataset into training, validation, and test sets, ensuring that the classes of diseases included in each set do not overlap. The reason we split the data without overlapping disease classes is to increase the usability of the ADR signal detection model. 
It is also because only a few classes of diseases exist in our dataset, and therefore there could be a data leakage if the same disease class exists in both training and validation. The class of disease means the classification up to the third digit of ICD-10 codes. Note that we make the inference very difficult by not letting the model know which classes of diseases are linked with drugs as ADRs. We use 80\% of data for training, 10\% for validation, and the remaining 10\% for testing. \\
To control the sparsity of a graph, we build two types of graphs where the edge-forming threshold is either low or high. When the edge-forming threshold is low, the graph has more edges, having more information as a result. We examine whether it is beneficial or detrimental to have more edge information. We distinguish two graphs by setting the thresholds for $\mathcal{E}_{\text{drug}}$ differently.  
The summary statistics of the constructed graphs and datasets are provided in Table~\ref{table:statistics}.

\begin{table}[t]
\centering
\small
\caption{
Model AUROC and AUPRC performances (including 95\%\ CI)
}
\begin{tabular}{L{0.25\columnwidth}C{0.35\columnwidth}C{0.35\columnwidth}}
\toprule
\textbf{Model} & \textbf{AUROC} & \textbf{AUPRC} \Tstrut \\ 
\midrule
LR & {0.631 $\pm$ 0.006} & 0.585 $\pm$ 0.007\Tstrut \\ 
NN & {0.739 $\pm$ 0.005} & 0.701 $\pm$ 0.006 \\
\midrule
$\text{{GCN}}_{\text{{\textit{low}}}}$ & \textbf{0.795} $\pm$ 0.006 & \textbf{0.775} $\pm$ 0.006 \\
$\text{{GAT}}_{\text{{\textit{low}}}}$ & 0.732 $\pm$ 0.005 & 0.686 $\pm$ 0.009\\
$\text{{adrGCN}}_{\text{{\textit{low}}}}$  & 0.755 $\pm$ 0.008 & 0.726 $\pm$ 0.009\\
$\text{{GCN}}_{\text{{\textit{high}}}}$ &  0.784 $\pm$ 0.006 & 0.761 $\pm$ 0.008 \\
$\text{{GAT}}_{\text{{\textit{high}}}}$ & 0.733 $\pm$ 0.008 & 0.692 $\pm$ 0.009\\
$\text{{adrGCN}}_{\text{{\textit{high}}}}$  & 0.756 $\pm$ 0.004 & 0.732 $\pm$ 0.006\\
\bottomrule
\end{tabular}
\label{table:experiments}
\end{table}

\subsection{Evaluations}
To verify the performance of the GCN-based approach, we compare GCN-based models with non-graph-based ML techniques. We apply vanilla GCN and its variants to examine the effect of considering the edge types. The followings are the models used for the graph embedding learnings. All the neural-network based models use two layers with a hidden dimension of 300.
\begin{itemize}
    \item
    \textbf{LR} is a logistic regression (LR) approach with information of the graph topology. The vector composed of initial node representations of the node itself and its neighbor nodes are input to the LR model. 
    The number of neighbors is limited to 10.
    \item 
    \textbf{NN} is a 2-layer fully-connected neural network which is solely based on the initial node representations. 
    \item
    $\text{\textbf{GCN}}_{\text{\textbf{\textit{low}}}}$
    is a graph convolution network, a representative GNN model that uses graph convolutions~\cite{kipf2016semi}.
    \item 
    $\text{\textbf{GAT}}_{\text{\textbf{\textit{low}}}}$ is a GNN that applies the attention mechanism on the node embeddings. Here we use GAT with two layers, where the number of heads are (4,4) for each layer.
    \item 
    $\text{\textbf{adrGCN}}_{\text{\textbf{\textit{low}}}}$ is an adapted version of GCN, that uses separate GCN layers according to the edge types and then aggregate them.  
    \item 
    $\textbf{GCN}_{\text{\textbf{\textit{high}}}}$,$\textbf{GAT}_{\text{\textbf{\textit{high}}}}$, $\textbf{adrGCN}_{\text{\textbf{\textit{high}}}}$ are the graph-based models applied to the sparser graph, i.e. the edge-forming threshold is high.
\end{itemize}

As shown in Table~\ref{table:experiments}, the proposed graph-based approaches surpass all the non-graph-based approaches. 
The best AUROC performance is achieved when \textbf{GCN} is applied with the low edge-forming threshold. The results show that the \textbf{GCN} model efficiently leverages the information from sufficiently selected edges. 
To see the robustness of the proposed method, we also examine whether our model works well for the infrequent drug-ADR pairs.
We evaluate model performance for the infrequent drug-ADR pairs, which are labeled as 'rare' or 'post-marketing' in SIDER.
As a result, the best average test accuracy in infrequent drug-ADR pairs is achieved with $\textbf{adrGCN}_{\text{\textbf{\textit{high}}}}$ (0.746), demonstrating that using multiple GCNs according to the edge types is useful to detect rare symptoms. According to the SIDER database, the ADRs with `rare` or `post-marketing` labels are reported with frequencies under 0.01.


\subsection{Newly-Described ADR Candidates}
\label{ssec:discovery}
To verify the power of the graph-based approach to discover ADR candidates which are unseen in the dataset, we extract the drug-disease pairs which are predicted to be positive with high probability --- over 0.97 but labeled as negative (false positive). To demonstrate the genuine power of graph-based methods, we exclude the candidates that are also positively predicted by the baseline neural network, which does not use relational information. As a result, clinical experts (M.D.) confirm that there exist pairs that are clearly considered to be real ADRs. The pairs are listed in Table \ref{table:discovery}.

Many of the discovered pairs, including umbrella terms like edema, are rather symptoms and signs than diseases. This can be explained by the fact that the SIDER database is less comprehensive to cover all the specific symptoms, that can be induced by taking medicine. Especially, cardiac murmur and abnormal reflex are frequent symptoms, but it is reasonable to say that the suggested pairs are ADRs. For example, Dasatinib is used to treat leukemia and can have significant cardiotoxicity, which can lead to cardiac murmurs. Hydroxycarbamide is a cytotoxic drug used for certain types of cancer, and it is known that cytotoxic medications can cause electrolyte imbalance leading to abnormal reflex.

There are also significant pairs such as alendronic acid and tetany in the third row. Severe and transient hypocalcemia is a well-known side-effect of bisphosphonates, which can lead to symptoms of tetany. Alendronic acid is classified as bisphosphonates, and therefore, tetany can be described as ADR of alendronic acid. Ibandronic acid and etidronic acid in the last two rows are also bisphosphonates, and the paired symptoms are relevant to the usage of bisphosphonates. Unspecified edema may signify bone marrow edema caused by bisphosphonate use, and electrolyte imbalance, which can lead to abnormal reflex, can be caused by etidronic acid use. All these explanations show that the ADR pairs we extract are based on various relations among drugs and diseases.

\begin{table}[t]
\centering
\small
\caption{
Newly-described drug-ADR pairs which are predicted, by the proposed method, to be positive with high probability.}
\begin{tabular}{C{0.35\columnwidth}C{0.35\columnwidth}C{0.2\columnwidth}}
\toprule
\textbf{Drug name}\Tstrut & \textbf{ADR symptom}\Tstrut & \textbf{Probability}\Tstrut \\
\midrule
Dasatinib &  Cardiac murmur & 0.985\\ 
Hydroxycarbamide & Abnormal reflex & 0.981 \\ 
Alendronic acid & Tetany & 0.978 \\  
Ibandronic acid & Unspecified edema & 0.976 \\  
Etidronic acid & Abnormal reflex & 0.972 \\ 
\bottomrule
\end{tabular}
\label{table:discovery}
\end{table}

\section{Conclusion}
\label{sec:conclusion}
In this study, we propose a novel graph-based approach for ADR detection by constructing a graph from the large-scale healthcare claims data. Our model can capture various relations among drugs and diseases, showing improved performance in predicting drug-ADR pairs. Furthermore, our model even predicts drug-ADR pairs that do not exist in the established ADR database, showing that it is capable of supplementing the ADR database. The explanation by clinical experts verifies that the graph-based method is valid for ADR detection. In this study, we only make inferences within the labeled dataset, yet we plan to make inferences on unlabeled data to discover unknown ADR pairs, which will be a huge breakthrough in ADR detection.
\paragraph{\textbf{Acknowledgements}}
K. Jung is with Automation and Systems Research
Institute (ASRI), Seoul National University. This work
was supported by the National Research Foundation of
Korea (NRF) grant funded by the Korea government (No. 2016R1A2B2009759).
\bibliographystyle{splncs04}
\bibliography{pakdd20}


\begin{appendix}

\section{Healthcare claims dataset}
For this study, we obtain data from the Sample Cohort Database (NHIS-NSC), a healthcare claims data established by national health insurance (NHI) of South Korea.
The NHIS-NSC is a retrospective cohort data from a population of one million patients sampled from 2002 to 2013, providing longitudinal observations of patient’s diagnosis, medication prescription, and procedures.
With this data, we extract target drugs and diseases and compute the statistics between any pairs of the drug-disease combinations. These statistics are used in determining edges in the drug-disease graph.

We represent drugs and diseases in NHIS-NSC data in a form of ATC codes (medication codes) and ICD-10 codes (International Classification of Diseases, 10th revision).
The number of converted ATC and ICD-10 codes are 1,201 and 1,872, respectively. 

\section{Adverse drug reaction dataset}
As a labeled dataset, we use Side Effect Resource (SIDER) database which contains 139,756 drug-side effect pairs over 1,430 drugs and 5,868 side effects. These were extracted from public information on recorded adverse drug reactions using natural language processing techniques.
To leverage medical and pharmacological knowledge, we extract drug and side effect information in a form of categorical codes (i.e. ATC and ICD-10 codes) with hierarchical structures. Since drug and side effect information in SIDER are represented as STITCH (Search Tool for Interactions of CHemicals) compound identifiers and UMLS (Unified Medical Language System) concept identifiers, we convert them to ATC and ICD-10 codes. The number of converted ATC and ICD-10 codes are 562 and 931, respectively.



\end{appendix}

%





\end{document}